\UseRawInputEncoding
\pdfoutput=1
\documentclass[letterpaper]{article} 
\usepackage[]{aaai23}  
\usepackage{times}  
\usepackage{helvet}  
\usepackage{courier}  
\usepackage[hyphens]{url}  
\usepackage{graphicx} 
\usepackage{natbib}  
\usepackage{caption} 
\usepackage{algorithm}
\usepackage{algorithmic}
\usepackage{amsfonts}
\usepackage{newfloat}
\usepackage{listings}
\usepackage{booktabs}
\usepackage{multirow}
\usepackage{bigstrut}
\usepackage{array}
\usepackage{amsmath}
\DeclareCaptionStyle{ruled}{labelfont=normalfont,labelsep=colon,strut=off} 
\floatstyle{ruled}
\newfloat{listing}{tb}{lst}{}
\floatname{listing}{Listing}
\title{Grasping Core Rules of Time Series through Pure Models}

\author {
	Gedi Liu,\textsuperscript{\rm 1,2,3,4}
	Yifeng Jiang, \textsuperscript{\rm 1,2,3,4}
	Yi Ouyang \textsuperscript{\rm 1,2,3,4}
	Keyang Zhong \textsuperscript{\rm 1,2,3,4}
	Yang Wang \textsuperscript{\rm 1,2,3,4}
}
\affiliations {
	\textsuperscript{\rm 1} National Innovation Center for Digital Fishery, China Agricultural University, Beijing, China\\
	\textsuperscript{\rm 2} Key Laboratory of Smart Farming Technologies for Aquatic Animal and Livestock, 
	\\Ministry of Agriculture and Rural Affairs, China Agricultural University, Beijing, China\\
	\textsuperscript{\rm 3} Beijing Engineering and Technology Research Center for Internet of Things in Agriculture, 
	\\China Agricultural University, Beijing, China\\
	\textsuperscript{\rm 4} College of Information and Electrical Engineering, China Agricultural University, Beijing, China\\
	\{2020308130522, 2020308130422, 2020309080216, 2020308130322, andy\_yangwang\}@cau.edu.cn
}

\begin{document}

\maketitle

\begin{abstract}
Time series underwent the transition from statistics to deep learning, as did many other machine learning fields. Although it appears that the accuracy has been increasing as the model is updated in a number of publicly available datasets, it typically only increases the scale by several times in exchange for a slight difference in accuracy. Through this experiment, we point out a different line of thinking, time series, especially long-term forecasting, may differ from other fields. It is not necessary to use extensive and complex models to grasp all aspects of time series, but to use pure models to grasp the core rules of time series changes. With this simple but effective idea, we created PureTS, a network with three pure linear layers that achieved state-of-the-art in 80\% of the long sequence prediction tasks while being nearly the lightest model and having the fastest running speed. On this basis, we discuss the potential of pure linear layers in both phenomena and essence.
The ability to understand the core law contributes to the high precision of long-distance prediction, and reasonable fluctuation prevents it from distorting the curve in multi-step prediction like mainstream deep learning models, which is summarized as a pure linear neural network that avoids over-fluctuating. Finally, we suggest the fundamental design standards for lightweight long-step time series tasks: input and output should try to have the same dimension, and the structure avoids fragmentation and complex operations.

\end{abstract}

\section{Introduction}
Time series refers to the regression task of predicting the future using data from the past. They are frequently employed in disciplines like finance~\cite{c:1,c:2} and weather forecasting~\cite{c:3,c:4} due to the pervasiveness of forecasting activities. Like other domains, time series have undergone a transition from statistical models based on hand-crafted features to deep learning models with stacked feature function blocks. The connectionist hypothesis~\cite{c:5}, along with a well-known corollary~\cite{hornik1991approximation} that multilayer perceptrons with activation layers can accommodate arbitrary dimensions and spaces, played a significant role in the development of machine learning and deep neural networks. This tide has essentially coupled the path of time series models' development with the trend of deep learning. Early statistics-based models are highly interpretable, ranging from the conventional autoregressive~\cite{c:7} and Gaussian distribution~\cite{c:8,c:9} fitting to the broader model of machine learning such as principal component analysis~\cite{c:10} and support vector machine regression~\cite{c:11}. Time series analysis is joining the ranks as a result of deep learning's influence in other domains. The recurrent neural network (RNN)~\cite{rumelhart1986learning} expands the Long Short Term Memory (LSTM)~\cite{c:13} and adds memory modules to time series; the success of convolutional neural networks in computer vision applies one-dimensional convolution to temporal information; and the attention mechanism~\cite{c:14}, as a delicate network structure, mimics human attention. Additionally, time series introduces graph neural networks as a new model structure, a slightly sparsely connected type that considers everything to be a graph structure. 

The above deep learning models add more functional modules or modify the training process to improve accuracy, which is representative of the trend of self-supervised or unsupervised learning, such as GPT-3~\cite{c:16} in Natural Language Processing (NLP) or MAE~\cite{he2022masked} in computer vision.
This trend motivates the unsupervised training of the time series model TS2Vec~\cite{yue2022ts2vec}, which learns the hierarchical structure and representation information of time series data, and then enhances the task performance through fine-tuning. They are partly reasonable because the time series also reflect changes in real-world data and may not be less complex than other AI tasks like image classification and NLP. Consider the several factors that must be carefully considered when utilizing time series to forecast agricultural greenhouse climate, for example. A variety of influencing elements can be incorporated, including the geographical setting in which the agricultural greenhouse is located, seasonal factors, and even human activities, such as the operation of the heater and humidifier or the gas flow brought on by ventilation. Unfortunately, despite adding additional parameters and lengthening the inference process, the model developed using this concept in the experimental chapter did not greatly enhance the model. Therefore, we reconsider and suggest that we do not need a big model like GPT-3 to duplicate every aspect of the environment but to understand the core rules in time series.

The recent works indicate that multilayer perceptrons have considerable potential due to their lightweight and purity. For instance, NBEATSx~\cite{c:17}, a multilayer perceptron capable of residual connections, produces state-of-the-art results on the M4 dataset~\cite{c:18}, and LightTS~\cite{zhang2022less} uses a downsampling method with the fundamental variation laws of nonlinear and periodic information to further improve the accuracy. Thence we accept the idea that MLP-based structures are particularly effective in identifying historical time series trends. Even if the aforementioned model is already relatively lightweight, it raises the question: Can multilayer perceptrons' potential still be explored? Is there a purer pattern out there that we have yet to find?

Hornik's theory~\cite{hornik1991approximation,hornik1989multilayer} shows that multilayer perceptrons with activation layers have the potential to fit any function, and constant activation layers make the fitting space clustered in narrow regions. The above narrow space is assumed to be a common law of time series, and a neural network with no activation layer is used to test the real data set and get amazing test results. In particular, a three-layer linear neural network, called PureTS, has produced state-of-the-art outcomes in 80\% of the prediction tasks for the long-sequence prediction dataset and is comparable to large-scale deep learning models in short sequence prediction. 
In order to explore the different operation rules of PureTS and other models, we start with the basic trigonometric functions and conclude that when the prediction step size increases, PureTS can learn trend information. Additionally, tests using both real data sets and trigonometric functions demonstrate that PureTS, which lacks complex nonlinearity, does not produce nearly as irregular polylines as other models, which is defined as avoiding over-fluctuating. PureTS is intuitively understood as the weight distribution of a given time step on the predicted data, similar to autoregressive models, and MLP-based models can also be regarded as nonlinear autoregressive models, which is similar to related work~\cite{elsayed2021we}. That shows the time series needs to pay attention to traditional models such as autoregressive models.

In this paper, we provide the following contributions:
\begin{itemize}
\item For the first time, it is suggested through an analysis of time series development that time series should grasp the core rules of data change rather than overfit the actual environment.
\item Our model PureTS is a simple and efficient approach that accomplishes state-of-the-art on the 80\% long step prediction problem.
\item The experiments indicate the linear layer's capacity to recognize data trends. Additionally, for the first time, we defined over-fluctuating, and analyzed the over-fluctuating phenomenon in the curve predicted by each model.
\end{itemize}

\section{Related Works}

\subsection{Statistical Methods}
The time series approach based on statistics is its traditional direction, and its predicting outcomes rely heavily on past data. ARIMA~\cite{c:7} (Auto-Regressive Integrated Moving Averages) is based on the following assumptions: The time series is stationary, therefore the researcher should use the logarithmic transformation or difference to make it such that the mean and variance do not fluctuate over time.
Holt-Winters~\cite{c:21}, an exponential smoothing method, consists of a prediction equation and three smoothing equations that fit level information, trend information, and periodic information, respectively. Gaussian processes~\cite{c:8,c:9} built on Bayesian techniques mimic the distribution of multivariate time series on continuous functions. The advantage of statistically based time series models is interpretability, while the drawback is adaptability. However, recent work has shown that statistical models are not inferior in accuracy, and autoregressive-based GBRT~\cite{elsayed2021we} outperforms a number of deep learning models.
\subsection{Deep Learning Methods}

Deep neural networks have evolved into numerous modules with specialized functionalities and sparse or dense neuron connections based on connectionism. It has been developed further to include LSTM~\cite{c:13} and GRU~\cite{c:22}, with memory blocks that govern whether to remember or forget, from the early multilayer perceptron to the RNN-based memory module. The primary function being investigated at this point is the time series model's capacity for long-term memory. Convolutional neural networks are adept at evaluating data fluctuations, and reasoning has emerged as a new area of research, as seen in SCINet~\cite{c:23}. The following development in the time series was caused by the attention mechanism. With the help of Informer~\cite{c:24}, DA-RNN~\cite{c:25}, the precision of the attention mechanism was improved. Graph neural networks are new models that convert dense connections to sparse graph structures, fitting information in explicit or implicitly graph-structured data, such as MTGNN~\cite{c:26}.

\begin{figure*}[t]
	\centering
	\includegraphics[width=1.8\columnwidth]{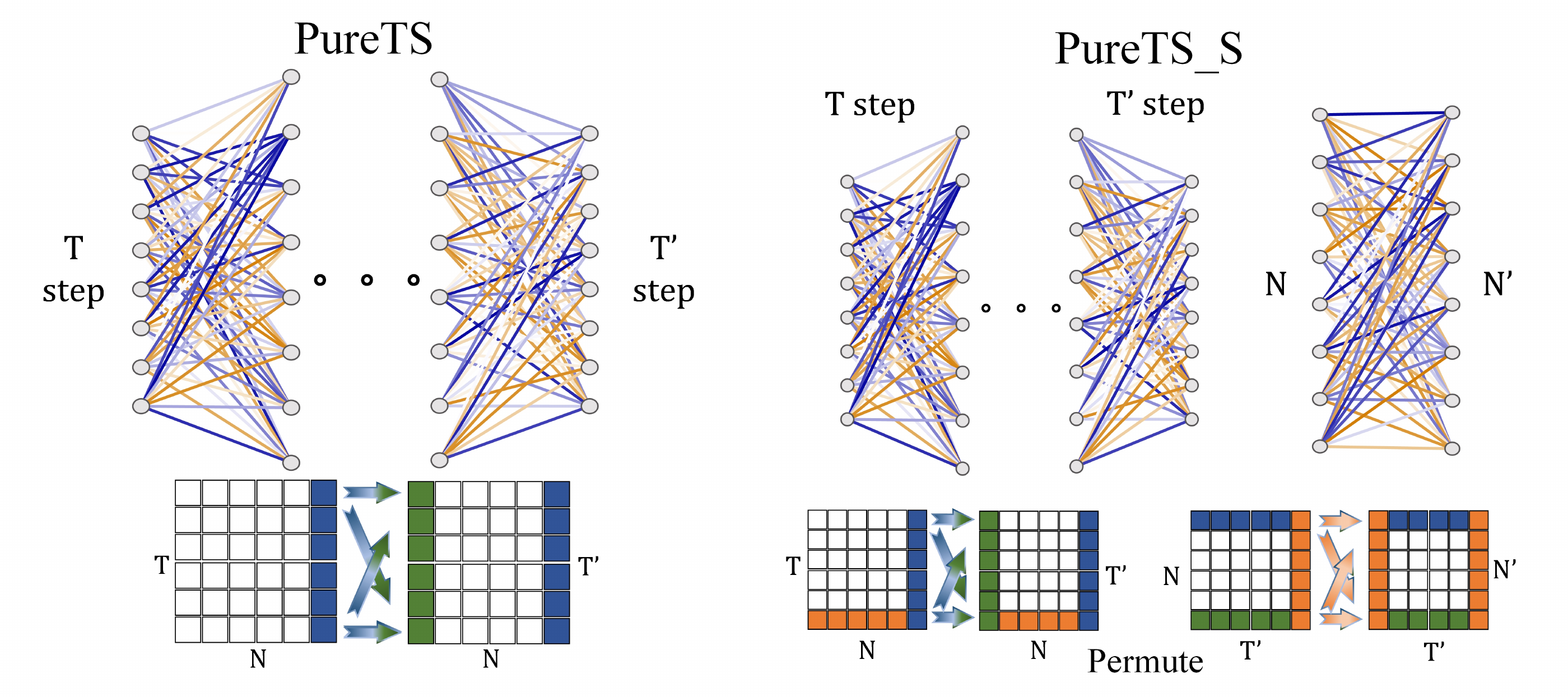} 
	\caption{Frame diagram of PureTS and PureTS\_S.}.
	\label{fig1}
\end{figure*}

\subsection{MLP-Based Models}

Despite being a part of deep learning, the MLP-based model can be seen as a new, lightweight time series product and should be segregated individually. It comes from the recent re-examination of multilayer perceptrons using deep learning, such as MLP-Mixer~\cite{c:27} for computer vision. Its influence has led to similar investigations in the realm of time series. On the M4 dataset, NBEATSx~\cite{c:17}, which has residual connections, can learn multi-level information and produce state-of-the-art outcomes. It has an impact on the lightweight model LightTS~\cite{zhang2022less}, adopts two downsampling techniques (including interval sampling and continuous sampling), and adds a new record to the M4 dataset. Through the aforementioned MLP-based lightweight time series model, we can see those multilayer perceptrons are sufficient to understand nonlinear and periodic information that occurs in the actual world.

\section{PureTS: Pure Linear Layer Model}
\subsection{Definition of Time Series}

The time series is a collection of data points that are abstracted mathematically from real-world time data and are arranged chronologically. The prediction task of time series is defined as: For a given time value $t$, the value corresponding to the current moment is $x_{t}$. The input length (window size) of the time series is $T$, and the input value is $X_{t}=\left\{ x_{t-T+1},x_{t-T+2},\ldots ,x_{t}| x_{i}\in \mathbb{R} ^{N} \right\}$, where N is the dimension of the corresponding value at a time step; its output length (horizon) is $T'$, the output value is  $X_{t}'=\left\{ x_{t+1},x_{t+2},\ldots ,x_{t+T'}| x_{i}\in \mathbb{R} ^{N}\right\}$. The above is called T'step prediction task, when $T'=1$, it is called single-step prediction, and when $T$ gradually increases to a certain extent, it is called long-sequence forecasting. 

\subsection{Details of the PureTS}
Hornik demonstrates that the neural network and activation layer has a great fitting ability, as long as there are enough hidden units, any Borel measurable function may be approximated to arbitrary precision from one finite-dimensional space to another finite-dimensional space~\cite{hornik1991approximation}. However, such a fitting capacity is a double-edged sword. It will lead to over-fitting and reduce the test accuracy. In order to fit the core rules between the input and output, we deleted the activation layer and employed the linear layer's projection ability. Time series often have two dimensions of information that should be fitted, spatiotemporal correlation. Therefore we classify PureTS into two primary examples based on this. As depicted in Figure~\ref{fig1}, one is a multilayer perceptron with only linear layers, and the other is a multilayer perceptron including spatiotemporal transformations.

PureTS is a non-trick linear neural network model (no activation layers, regularizers, or dropout). Basic PureTS is depicted in the figure on the left. The number of linear layers in this experiment is three(adjustable), depending on the data's intricacy. One linear layer is usually enough to fit the overall trend of the time series in short-step forecasting. Enter the vector $X_{t}=\left\{ x_{t-T+1},x_{t-T+2},\ldots ,x_{t}| x_{i}\in \mathbb{R} ^{N} \right\}$ to perform the "Permute" operation, transfer the time dimension to the projectable $X_{N}=\left\{ x_{1},x_{2},\ldots ,x_{N}| x_{i}\in \mathbb{R} ^{T} \right\}$, perform linear projections of several time dimensions to get $X_{N}'=\left\{ x_{1}',x_{2}',\ldots ,x_{N}'| x_{i}'\in \mathbb{R} ^{T'} \right\}$, and finally, perform the "Permute" operation to obtain the predicted value $X_{t}'=\left\{ x_{t+1},x_{t+2},\ldots ,x_{t+T'}| x_{i}\in \mathbb{R} ^{N} \right\}$. In the image on the right, PureTS\_S has spatial projection applied, and the feature dimension has been a linear transformation to produce $X_{t}''=\left\{ x_{t+1}',x_{t+2}',\ldots ,x_{t+T'}'| x_{i}'\in \mathbb{R} ^{N'} \right\}$.

\section{Experiment}

\begin{table*}[htbp]
	\centering
	\begin{tabular}{|c|c|cc|cc|cc|cc|cc|}
		\hline
		\multirow{3}[6]{*}{Methods} & \multirow{3}[6]{*}{Metrics} & \multicolumn{2}{c|}{ETTh1} & \multicolumn{2}{c|}{ETTh2 } & \multicolumn{2}{c|}{ETTm1} & \multicolumn{2}{c|}{Weather} & \multicolumn{2}{c|}{Electricity} \bigstrut\\
		\cline{3-12}          &       & \multicolumn{2}{c|}{horizon} & \multicolumn{2}{c|}{horizon} & \multicolumn{2}{c|}{horizon} & \multicolumn{2}{c|}{horizon} & \multicolumn{2}{c|}{horizon} \bigstrut\\
		\cline{3-12}          &       & 336   & 720   & 336   & 720   & 288   & 672   & 336   & 720   & 720   & 960 \bigstrut\\
		\hline
		\multirow{2}[2]{*}{LogTrans} & MSE   & 0.942  & 1.109  & 3.711  & 2.817  & 1.728  & 1.865  & 0.666  & 0.741  & 0.311  & 0.333  \bigstrut[t]\\
		& MAE   & 0.766  & 0.843  & 1.587  & 1.356  & 1.656  & 1.721  & 0.584  & 0.611  & 0.397  & 0.413  \bigstrut[b]\\
		\hline
		\multirow{2}[2]{*}{Reformer} & MSE   & 1.919  & 2.177  & 3.798  & 5.111  & 1.632  & 1.943  & 1.770  & 2.548  & 1.883  & 1.973  \bigstrut[t]\\
		& MAE   & 1.090  & 1.218  & 1.508  & 1.793  & 0.886  & 1.006  & 0.997  & 1.407  & 1.002  & 1.185  \bigstrut[b]\\
		\hline
		\multirow{2}[2]{*}{Informer} & MSE   & 0.884  & 0.941  & 1.665  & 2.340  & 1.219  & 1.651  & 0.623  & 0.685  & 0.308  & 0.328  \bigstrut[t]\\
		& MAE   & 0.753  & 0.768  & 1.035  & 1.209  & 0.871  & 1.002  & 0.546  & 0.575  & 0.385  & 0.406  \bigstrut[b]\\
		\hline
		\multirow{2}[2]{*}{Autoformer} & MSE   & 0.724  & 0.898  & 1.386  & 2.445  & 0.575  & 0.599  & \textbf{0.492 } & \textbf{0.527 } & 0.259  & 0.291  \bigstrut[t]\\
		& MAE   & 0.651  & 0.743  & 0.892  & 1.226  & 0.527  & 0.542  & \textbf{0.491 } & \textbf{0.503 } & 0.361  & 0.381  \bigstrut[b]\\
		\hline
		\multirow{2}[2]{*}{LSTMa} & MSE   & 1.152  & 1.682  & 3.276  & 3.711  & 1.598  & 2.530  & 1.497  & 1.314  & 1.528  & 1.343  \bigstrut[t]\\
		& MAE   & 0.794  & 1.018  & 1.375  & 1.520  & 0.952  & 1.259  & 0.889  & 0.875  & 0.945  & 0.886  \bigstrut[b]\\
		\hline
		\multirow{2}[2]{*}{LSTNet} & MSE   & 2.477  & 1.925  & 1.372  & 2.403  & 1.009  & 1.681  & 0.714  & 0.773  & 0.442  & 0.473  \bigstrut[t]\\
		& MAE   & 1.193  & 1.084  & 2.429  & 3.403  & 1.902  & 2.701  & 0.607  & 0.643  & 0.433  & 0.443  \bigstrut[b]\\
		\hline
		\multirow{2}[2]{*}{SCINet} & MSE   & 0.528  & 0.597  & 0.657  & 1.118  & 0.350  & 1.214  & 0.540  & 0.577  & 0.234  & 0.272  \bigstrut[t]\\
		& MAE   & 0.513  & 0.571  & 0.576  & 0.776  & 0.405  & 0.836  & 0.521  & 0.549  & 0.332  & 0.361  \bigstrut[b]\\
		\hline
		\multirow{2}[2]{*}{LightTS} & MSE   & \textit{0.466 } & \textit{0.542 } & \textit{0.497 } & \textit{0.739 } & \textit{0.272 } & \textit{0.391 } & \textit{0.527 } & \textit{0.554 } & \textit{0.219 } & \textit{0.235 } \bigstrut[t]\\
		& MAE   & \textit{0.468 } & \textit{0.536 } & \textit{0.499 } & \textit{0.610 } & \textit{0.335 } & \textit{0.420 } & \textit{0.509 } & \textit{0.525 } & \textit{0.318 } & \textit{0.329 } \bigstrut[b]\\
		\hline
		\multirow{2}[2]{*}{PureTS} & MSE   & \textbf{0.438 } & \textbf{0.473 } & \textbf{0.360 } & \textbf{0.581 } & \textbf{0.245 } & \textbf{0.345 } & 0.561  & 0.614  & \textbf{0.196 } & \textbf{0.214} \bigstrut[t]\\
		& MAE   & \textbf{0.433 } & \textbf{0.492 } & \textbf{0.401 } & \textbf{0.530 } & \textbf{0.304 } & \textbf{0.376 } & 0.532  & 0.566  & \textbf{0.254 } & \textbf{0.302} \bigstrut[b]\\
		\hline
	\end{tabular}%
	\caption{Long sequence prediction task results. The model with the best result is in \textbf{bold}, the second result in \textit{italics}. The experimental results except PureTS are from~\cite{zhang2022less}.}
	\label{table1}%
\end{table*}%
\begin{table*}[htbp]
	\centering
	\small
	\begin{tabular}{|c|c|cccc|cccc|}
		\hline
		\multicolumn{1}{|c|}{\multirow{3}[6]{*}{Methods}} & \multirow{3}[6]{*}{Metrics} & \multicolumn{4}{c|}{Exchange-Rate} & \multicolumn{4}{c|}{Solar-Energy} \bigstrut\\
		\cline{3-10}          & \multicolumn{1}{c|}{} & \multicolumn{4}{c|}{horizon} & \multicolumn{4}{c|}{horizon} \bigstrut\\
		\cline{3-10}          & \multicolumn{1}{c|}{} & 3     & 6     & 12    & 24    & 3     & 6     & 12    & 24 \bigstrut\\
		\hline
		\multicolumn{1}{|c|}{\multirow{2}[2]{*}{AR}} & RSE   & 0.0228 & 0.0279 & 0.0353 & 0.0445 & 0.2435 & 0.3790 & 0.5911 & 0.8699 \bigstrut[t]\\
		& CORR  & 0.9734 & 0.9656 & 0.9526 & 0.9357 & 0.9710 & 0.9263 & 0.8107 & 0.5314 \bigstrut[b]\\
		\hline
		\multicolumn{1}{|c|}{\multirow{2}[2]{*}{VARMLP}} & RSE   & 0.0265 & 0.0394 & 0.0407 & 0.0578 & 0.1922 & 0.2679 & 0.4244 & 0.6841 \bigstrut[t]\\
		& CORR  & 0.8609 & 0.8725 & 0.8280 & 0.7675 & 0.9829 & 0.9655 & 0.9058 & 0.7149 \bigstrut[b]\\
		\hline
		\multicolumn{1}{|c|}{\multirow{2}[2]{*}{GP}} & RSE   & 0.0239 & 0.0272 & 0.0394 & 0.0580 & 0.2259 & 0.3286 & 0.5200 & 0.7973 \bigstrut[t]\\
		& CORR  & 0.8713 & 0.8193 & 0.8484 & 0.8278 & 0.9751 & 0.9448 & 0.8518 & 0.5971 \bigstrut[b]\\
		\hline
		\multicolumn{1}{|c|}{\multirow{2}[2]{*}{RNN-GRU}} & RSE   & 0.0192 & 0.0264 & 0.0408 & 0.0626 & 0.1932 & 0.2628 & 0.4163 & 0.4852 \bigstrut[t]\\
		& CORR  & 0.9786 & \textit{0.9712} & 0.9531 & 0.9223 & 0.9823 & 0.9675 & 0.9150 & 0.8823 \bigstrut[b]\\
		\hline
		\multicolumn{1}{|c|}{\multirow{2}[2]{*}{TCN}} & RSE   & 0.0217 & 0.0263 & 0.0393 & 0.0492 & 0.1940 & 0.2581 & 0.3512 & 0.4732 \bigstrut[t]\\
		& CORR  & 0.9693 & 0.9633 & 0.9531 & 0.9223 & 0.9835 & 0.9602 & 0.9321 & 0.8812 \bigstrut[b]\\
		\hline
		\multicolumn{1}{|c|}{\multirow{2}[2]{*}{MTGNN}} & RSE   & 0.0194 & 0.0259 & 0.0349 & 0.0456 & \textit{0.1778} & 0.2348 & 0.3109 & 0.4270 \bigstrut[t]\\
		& CORR  & 0.9786 & 0.9708 & \textit{0.9551} & \textit{0.9372} & \textit{0.9852} & 0.9726 & 0.9509 & 0.9031 \bigstrut[b]\\
		\hline
		\multicolumn{1}{|c|}{\multirow{2}[2]{*}{SCINet}} & RSE   & 0.0179 & 0.0249 & 0.0344 & 0.0462 & 0.1788 & \textit{0.2319} & \textit{0.3049} & \textit{0.4249} \bigstrut[t]\\
		& CORR  & 0.9744 & 0.9655 & 0.9493 & 0.9279 & 0.9849 & \textit{0.9735} & \textit{0.9529} & \textit{0.9026} \bigstrut[b]\\
		\hline
		\multicolumn{1}{|c|}{\multirow{2}[2]{*}{LightTS}} & RSE   & \textit{0.0178} & \textit{0.0246} & \textit{0.0339} & \textit{0.0453} & \textbf{0.1704} & \textbf{0.2212} & \textbf{0.2930} & \textbf{0.4133} \bigstrut[t]\\
		& CORR  & \textbf{0.9798} & 0.9710 & 0.9548 & 0.9360 & \textbf{0.9866} & \textbf{0.9761} & \textbf{0.9564} & \textbf{0.9065} \bigstrut[b]\\
		\hline
		\multicolumn{1}{|c|}{\multirow{2}[2]{*}{PureTS}} & RSE   & \textbf{0.0174} & \textbf{0.0242} & \textbf{0.0331} & \textbf{0.0429} & 0.2572 & 0.3211 & 0.4481 & 0.5597 \bigstrut[t]\\
		& CORR  & \textit{0.9793} & \textbf{0.9713} & \textbf{0.9571} & \textbf{0.9392} & 0.9682 & 0.9466 & 0.8904 & 0.8310 \bigstrut[b]\\
		\hline
		\multicolumn{1}{|c|}{\multirow{2}[2]{*}{PureTS\_S}} & RSE   & \textit{0.0178} & 0.0255 & 0.0356 & 0.0519 & 0.1945 & 0.2624 & 0.3606 & 0.5100 \bigstrut[t]\\
		& CORR  & 0.9734 & 0.9678 & 0.9540 & 0.9322 & 0.9820 & 0.9662 & 0.9323 & 0.8567 \bigstrut[b]\\
		\hline
	\end{tabular}%
	\caption{Short sequence prediction task results. The model with the best result is in \textbf{bold}, the second result in \textit{italics}. The experimental results except PureTS and PureTS\_S are from~\cite{zhang2022less}.}
	\label{table2}%
\end{table*}%

\begin{figure*}[htbp]
	\centering
	\includegraphics[width=1.8\columnwidth]{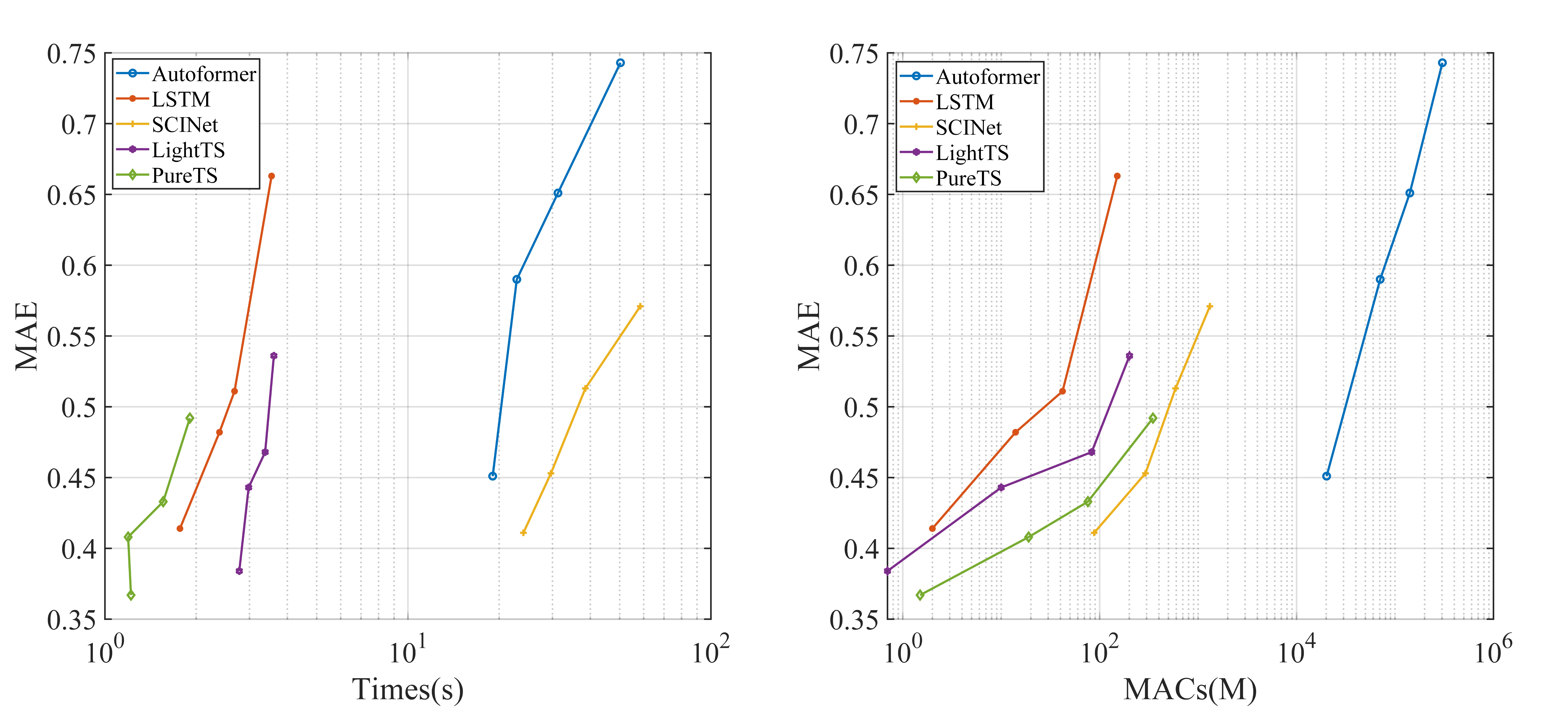} 
	\caption{Model MACs and runtime comparison. The different points represent different step length settings in the ETTh1 dataset, and the steps from bottom to top are 48, 168, 336, 720.}.
	\label{fig2}
\end{figure*}

In this chapter, we will analyze two datasets with contrasting characteristics and their extension tasks, validate the predictive capability of PureTS, and compare the running times and parameter sizes of several models horizontally. The eight publicly available benchmark datasets are split into short sequence prediction and long sequence prediction in accordance with the previous experimental settings~\cite{zhang2022less,c:24,c:33}. The long-sequence prediction data set adopts the evaluation indicators of the Mean Squared Error (MSE) and Mean Absolute Error (MAE), and the short-sequence data set adopts the evaluation indicators of the Root Relative Squared Error (RSE) and Empirical Correlation Coefficient (CORR).

The test model is classified into statistics and deep learning directions based on its parameters and computational complexity. The deep learning direction is further divided into four categories by additional functional modules.
\begin{itemize}
	\item Statistically-Based models: AR. GP~\cite{c:9}.
	
	\item Transformer-Based models: LogTrans~\cite{c:28}. Reformer~\cite{c:29}. Informer~\cite{c:24}. Autoformer~\cite{c:30}.
	
	\item CNN-Based models:
	TCN~\cite{c:31}. SCINet~\cite{c:23}.
	
	\item RNN-Based models:
	RNN-GRU~\cite{c:32}. LSTNet~\cite{c:33}. LSTMa~\cite{c:34}.
	
	\item GNN-Based models:
	MTGNN~\cite{c:26}.
	
	\item MLP-Based models:
	VARMLP~\cite{c:36}. LightTS~\cite{zhang2022less}. 
\end{itemize}

\subsection{Main Results Presentation}
In the long sequence prediction task, PureTS is set up as a three-layer multilayer perceptron. The comparison results are shown in Table~\ref{table2}.

The results show that PureTS achieves 80\% state-of-the-art with only three linear layers. MLP-based PureTS and LightTS generally outperform other models with special function blocks. According to LightTS statistics~\cite{zhang2022less}, for the longest prediction horizon, LightTS lowers MSE by 9.21\%, 33.90\%, 34.18\%, and 13.60\% on the ETTh1, ETTh2, ETTm1, and Electricity datasets, respectively. On this basis, the error of PureTS is lower than LightTS, respectively decreased by 8.94\%, 15.09\%, 11.70\%, and 8.94\%. Additionally, CNN-based models downsample with convolution kernels, whereas multilayer perceptrons downsample with structural refinement or pure linear maps like PureTS, but they can better represent long-term information. We also ran additional experiments with medium step size in appendix, PureTS can also achieve state-of-the-art. Even though PureTS excels at comprehending long-term trends, particularly the basic rules of time series, it is not bad at short-step forecasting.

Additionally, as indicated in Table~\ref{table2}, We conducted experiments on the short-step dataset. The Exchange-Rate data's feature dimension is 8. The result of PureTS is better than that of PureTS\_S, and the prediction result has been achieved state-of-the-art. In Solar-Energy, the feature dimension rises to 137, making prediction more difficult, and PureTS's outcome is less effective than PureTS\_S. This indicates that deep learning networks still require careful design for models with high feature dimensions.

\begin{figure*}[htbp]
	\centering
	\includegraphics[width=1.9\columnwidth]{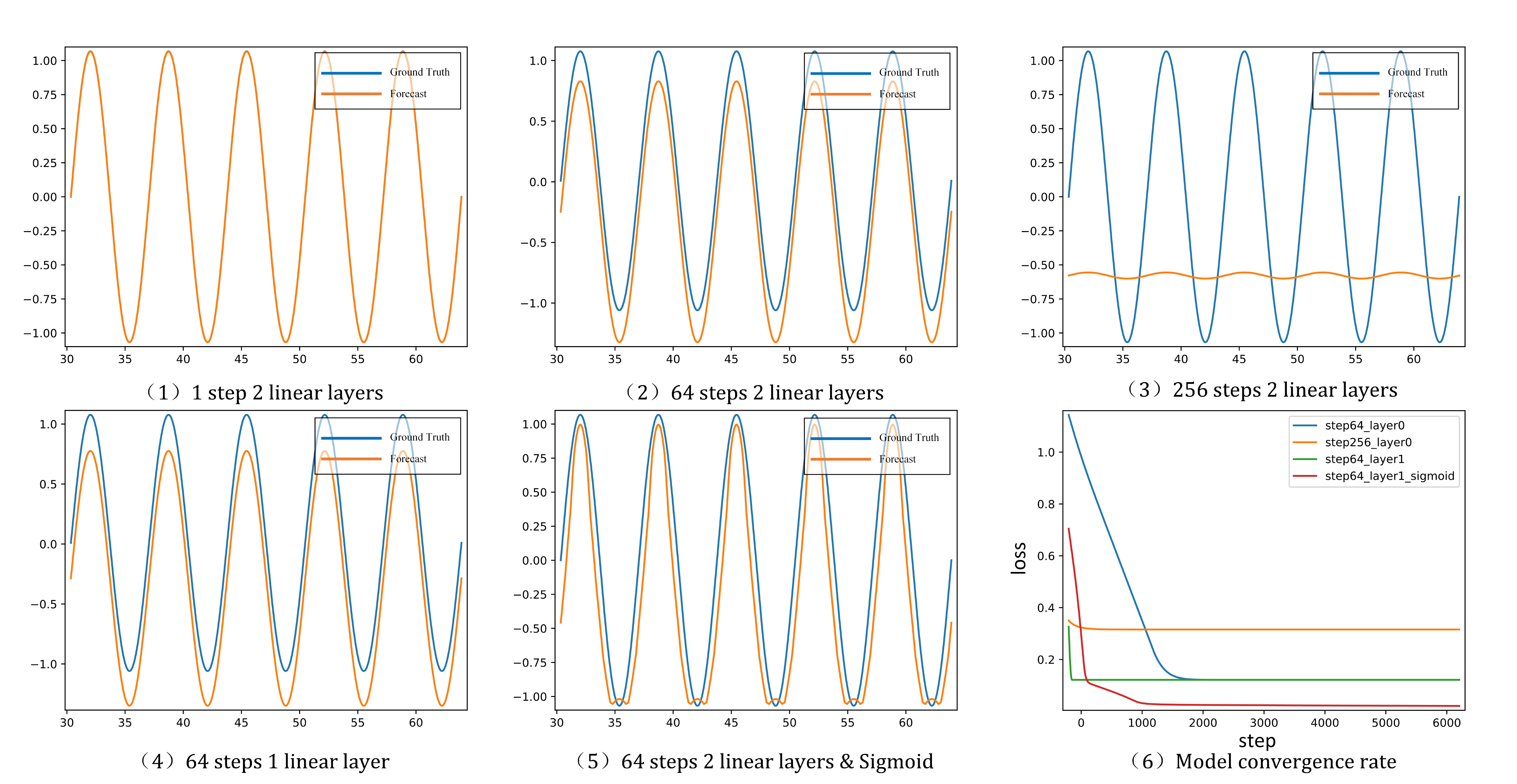} 
	\caption{Fitting effect of PureTS under different conditions.}.
	\label{fig3}
\end{figure*}

\subsection{High Speed and Lightweight}
For time series models, the model's calculation volume and running speed are also crucial evaluation factors. The dataset will choose ETTh1 for long-sequence prediction tasks once we compare the multiple-accumulate operations (MACs) and inference speed of the recent time series models Informer, SCINet, LSTM, LightTS, and PureTS, as illustrated in Figure~\ref{fig2}. 

Unquestionably, PureTS is the fastest model. It is interesting to note that the model's running speed is not directly correlated with the model's MACs or parameters, specifically in GPU inference, further optimization is easier with the pure model. The parameter comparison can be found in the appendix.  Following the ShuffleNet V2~\cite{ma2018shufflenet} design principle for lightweight models, time series designers can also refer to:
\begin{itemize}
	\item Equal channel width minimizes memory access cost and tries to keep the same input and output dimensions. 
	\item Network fragmentation reduces parallelism and purifies the model as much as possible. 
	\item Element operations must be considered, especially in lightweight models where prediction accuracy is frequently gained through fragmentation operations.
\end{itemize}

\section{Discussion}

Despite being quick and pure, PureTS is not far behind in forecasting tasks, particularly in capturing long-term trends. In this chapter, we will examine PureTS's fundamental performance and gather proof of the requirement for time series to understand the fundamental rules of change.

\subsection{Specific Prediction Space}
Previous research has shown that a multilayer perceptron with the activation layer has a dense prediction space, while a purely linear layer can only fit linear and constant terms of the input and has a restricted and focused prediction space. The experimental portion then demonstrates that pure linear layers can perform well in a variety of prediction tasks. As a result, we hypothesize that the majority of time series functions follow a common law, which is a constrained but representative function space that the linear layer can learn when the time series prediction step size increases.

\begin{figure*}[htpb]
	\centering
	\includegraphics[width=2.1\columnwidth]{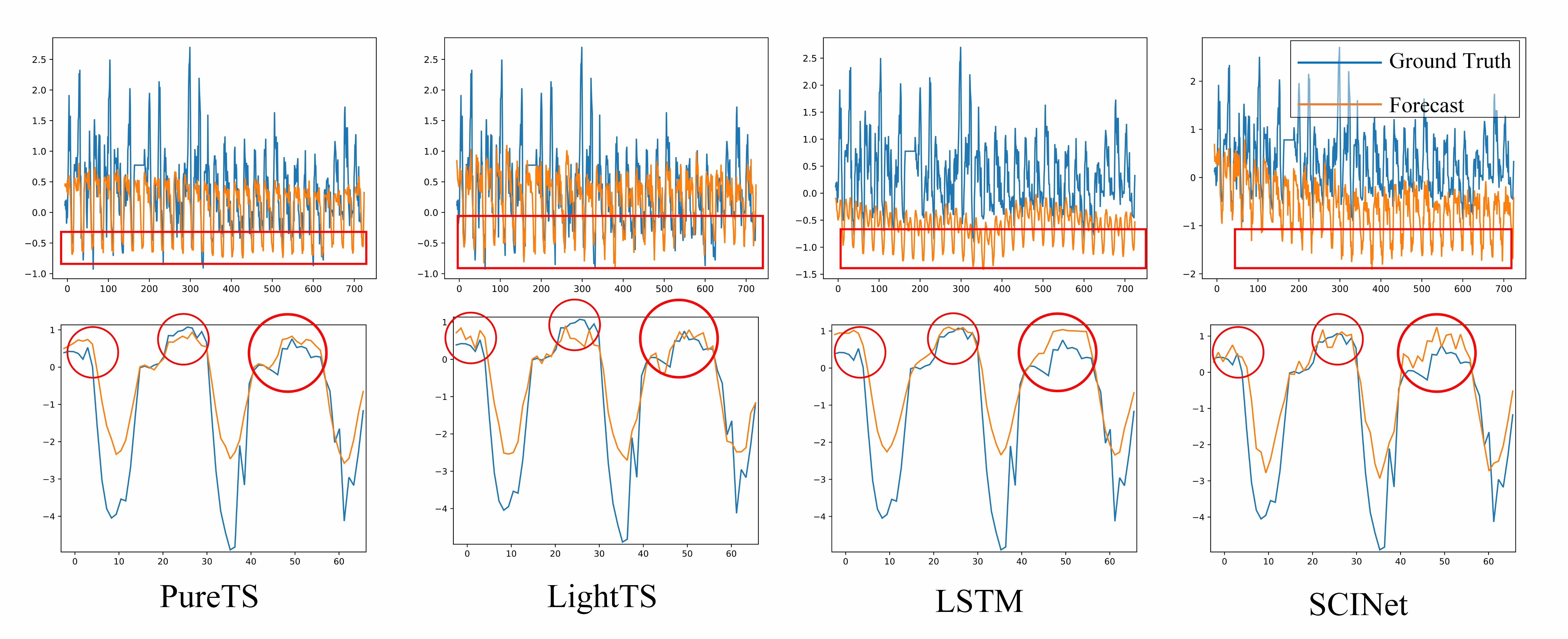} 
	\caption{The prediction results of the real data sets of each model are displayed. The upper part is a visualization of 720 steps, and the lower part is a detailed display of 64 steps.}.
	\label{fig5}
\end{figure*}

\subsection{Linear Layer Revisited}

It is presumable that the linear layer improves generalization while decreasing the model's capacity to fit higher-order terms. A Fourier series expansion can be applied to any periodic function $f\left( x\right)$, therefore we begin with the most fundamental $\sin x$ observations. The neural network is used to predict $t^{\prime}$-step $X'$ from t-step $X$. Using $\sin x$ as the prediction object of the time series.

The fitting state of $f\left( x\right)$ under five varying conditions is detailed in Figure~\ref{fig3}. Long-sequence time prediction is based on Figure (2), which depicts the outcome of a 64-step, two-layer linear layer fitting the $f\left( x\right)$ curve while sacrificing peak value fitting in favor of grasping the overall trend. Figure (1) reduces the step size to 1. The findings demonstrate that $f\left( x\right)$ can be perfectly fitted by the linear layer. It suggests that in an ideal scenario, a time series task could be predicted using linear layers. Based on the baseline, Figure (3) greatly raises the step size to 256. The results are in line with expectations: the image periodicity is well fitted, the peak fitting is abandoned, and the long-term series largely grasps the general trend of the time series information rather than the precise value. Figure (4) lowers the two neural network layers to one, and it can be concluded that the accuracy of the first layer and the two levels of the linear layer are the same, which is compatible with the theory we provided, because their output terms are the sum of the primary term and the constant term of the input term. However, the difference between them is in the fitting speed. The layer for Sigmoid activation is added in Figure (5). Activation layers boost peak fit while erasing the overall trend, making the curve abnormal. The abnormal phenomenon that the nonlinearity introduced makes the fitting curve appear additional fluctuation is defined as over-fluctuating. The convergence of the aforementioned five conditions is shown in Figure (6). To summarize, the larger the step size, the faster the model converges, and the accuracy drops, especially when the peak fitting is poor, which improves understanding of the overall rule of the model. Only the model's convergence condition is impacted by an increase in the linear layer. The peak accuracy of the model could increase with the addition of the activation layer, but nonlinearity will cause over-fluctuating, and the pace of convergence will slow down since particular values must be fitted.

\subsection{Trend Analysis and Over-fluctuating}
This section will analyze the model's ability to grasp the trend and the phenomenon of over-fluctuating by visualizing the four best-performing models.

First, examine the overall fit of each model. In the upper half of Figure~\ref{fig5}, in the case of long-step prediction fitting, all models show an excellent periodic fit. In terms of grasping trends, the MLP-based PureTS and LightTS models are better able to comprehend the general trend of the data, while LSTM and SCINet exhibit deviations in prediction. According to the details highlighted in the red box, PureTS and LSTM fit smoothly at the peak because their models contain fewer nonlinear components. However, the LightTS and SCINet models fit well at the peak and have larger fluctuations because they are intentionally designed to introduce complex nonlinearity.

Secondly, in the lower part of Figure~\ref{fig5}, some details of the long-step prediction are shown, which can also verify the conclusion of the Linear Layer Revisited section. The smoothest is LSTM, but it does not fit well. At the peak, PureTS added the proper fluctuations, but did not add more fluctuations to get too close to the peak. While the well-designed LightTS and SCINet can perform better at the peak, they inevitably introduce over-fluctuating. Similar to when fitting a trigonometric function, the distortion here is similar to losing the overall smoothness and periodicity in order to fit the peaks.
\section{Limitations}
Although PureTS achieves amazing results on real datasets, we still cannot theoretically explain why PureTS has such high accuracy, which is worth following up in subsequent work. We found that PureTS does not produce unreasonable polylines by refining the fitted curve because no nonlinearity is introduced, but we cannot explain their connection and why no nonlinearity can be close to the real curve.
\section{Conclusions}
We moved the emphasis of time series fitting to grasping its core variation laws, investigated pure linear neural networks, established PureTS, and achieved state-of-the-art results in 80\% of the tasks predicted by long-step datasets. Based on this finding, we suggest that the time series task is distinct from other fields, particularly in long series forecasting, where the model's primary objective is to understand the time series' fundamental trend rather than every particularity of the surrounding environment. Precision, convergence speed, and other angles have demonstrated PureTS's effectiveness. We define over-fluctuating based on an analysis of the fitted curve, and discover that excessive nonlinearity will make over-fluctuating result in distortion. On this foundation, we suggest several guidelines for establishing the lightweight model of time series long series prediction, including using the same dimension for input and output, preventing model operation fragmentation, and avoiding complicated calculations.
\section{Acknowledgments}
This work is supported by the National Key Research and Development Program of China: Sino-Malta Fund 2019 “Research and Demonstration of Real-time Accurate Monitoring System for Early-stage Fish in Recirculating Aquaculture System” (AquaDetector, Grant No. 2019YFE0103700), 2115 Talent Development Program of China Agricultural University, Major Science and Technology Innovation Fund 2019 of Shandong Province (Grant No. 2019JZZY010703), Overseas High-level Youth Talents Program (China Agricultural University, China, Grant No. 62339001). Special thanks to Yu Guo et al. 's help in the learning process.
\bibliographystyle{aaai23}
\bibliography{aaai23}

\clearpage
\section{Appendix}

\section{Model Parameter Statistics}
\begin{figure*}[t]
	\centering
	\includegraphics[width=2\columnwidth]{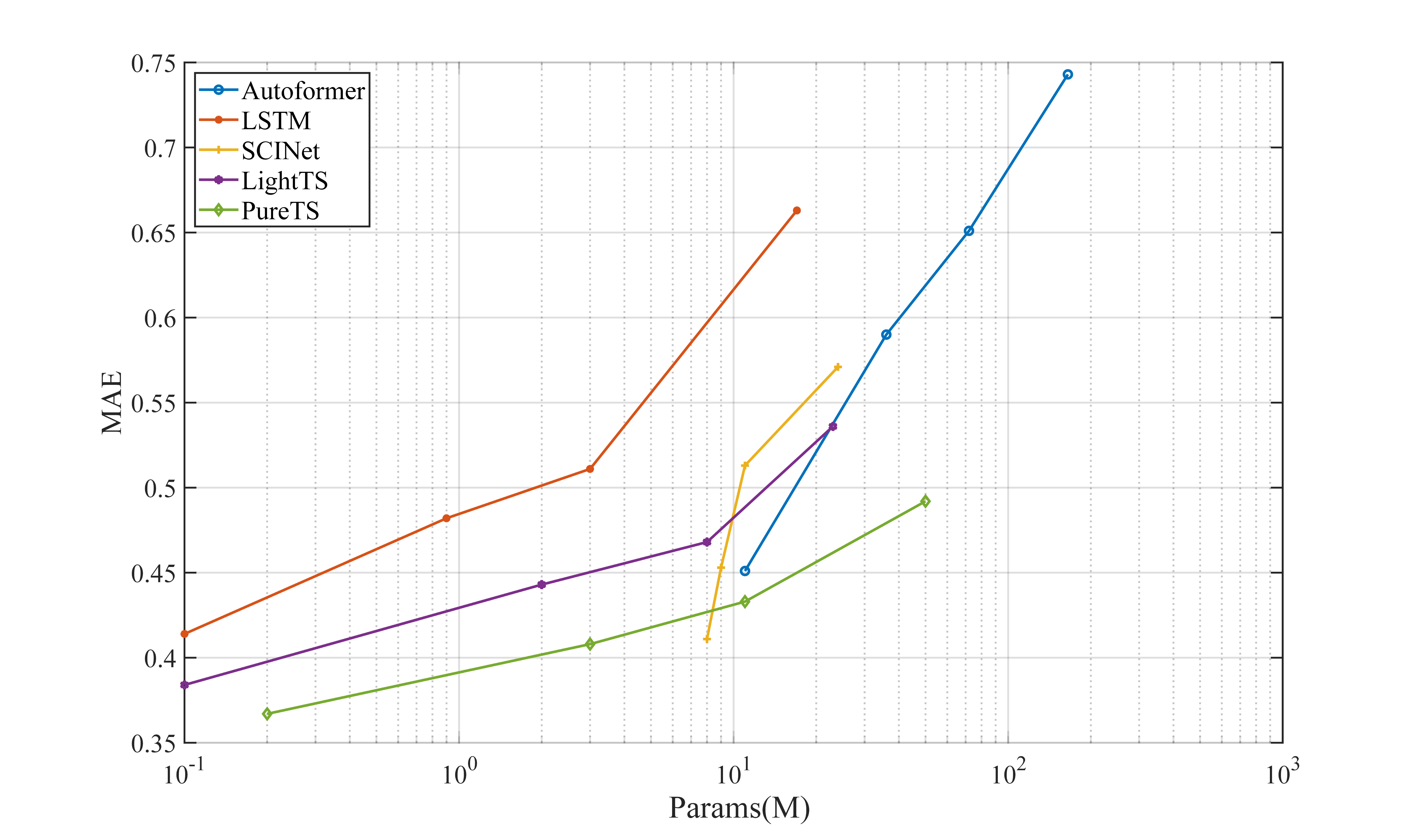} 
	\caption{The parameters and MAE statistics of each model. The different points represent different step length settings in the ETTh1 dataset, and the steps from bottom to top are 48, 168, 336, 720.}.
	\label{figure1}
\end{figure*}

In the ETTh1 long sequence prediction task, the statistics of the parameters of each model are shown in Figure~\ref{figure1}.

\section{Supplementary Results}
Results for short strides in long sequence datasets are shown in Table~\ref{tableadd2}.

Results in short sequence datasets are shown in Table~\ref{table3}.

\section{Dataset Details}
All data sets used in the experiment are shown in Table~\ref{dataset}. ETTh1, ETTh2, ETTm1, and Weather belong to long-sequence data sets, while Solar-Energy, Traffic, Electricity and Exchange-Rate belong to short-sequence data sets.
\begin{table}[htb]
	\begin{tabular}{|c|c|c|c|}
		\hline
		Datasets & Feature & Timesteps & Granularity \bigstrut\\
		\hline
		ETTh1 & 7     & 17420 & 1 hour \bigstrut\\
		\hline
		ETTh2 & 7     & 17420 & 1 hour \bigstrut\\
		\hline
		ETTm1 & 7     & 69680 & 15 minutes \bigstrut\\
		\hline
		Weather & 12    & 35064 & 1 hour \bigstrut\\
		\hline
		Solar-Energy & 137   & 17544 & 10 minutes \bigstrut\\
		\hline
		Traffic & 862   & 52560 & 1 hour \bigstrut\\
		\hline
		Electricity & 321   & 26304 & 1 hour \bigstrut\\
		\hline
		Exchange-Rate & 8     & 7588  & 1 day \bigstrut\\
		\hline
	\end{tabular}%
	\caption{The details of the experimental data set, the classification follows LightTS~\cite{zhang2022less}.}
	\label{dataset}%
\end{table}%
\begin{itemize}
	\item ETT (Electricity Transformer Temperature): The ETT is
	a crucial indicator in the electric power long-term deployment, collected by~\cite{c:24}. It was separated as ETTh1, ETTh2 for 1-hour-level and ETTm1 for 15-minute-level. The train/val/test is 12/4/4 months.
	\item Weather: Local climatological data for almost 1,600 U.S. locales are included in this collection for the four years between 2010 and 2013, with data points being gathered every hour. The train/val/test is 28/10/10 months.
	\item ECL (Electricity): The electricity dataset records 321 users' usage of electricity every 15 minutes over the course of two years (2012-2014). In long sequence forecasting, the train/val/test is 7/1/2. In short sequence forecasting, the train/val/test is 6/2/2.
	\item Solar-Energy: The 137 photovoltaic plants in Alabama State that produced solar energy in 2006 are tracked in this dataset every ten minutes. The train/val/test is 6/2/2.
	\item Traffic: This dataset compiles information on California's hourly road occupancy rates for the two years (2015–2016). 862 variables are contained in each timestamp. The train/val/test is 6/2/2.
	\item Exchange-Rate: The daily exchange rates of eight nations, including Australia, Canada, China, Japan, New Zealand, Singapore, Switzerland, and the United Kingdom, are compiled in this dataset from 1990 to 2016. The train/val/test is 6/2/2.
\end{itemize}

\section{Evaluation Metrics}
In the long series prediction, we employ Mean Squared Error (MSE) and Mean Absolute Error (MAE) as indicators and follow the setting~\cite{c:24}. The formula is as follows:
\begin{itemize}
	\item Mean Squared Error (MSE):
	$$
	M S E=\frac{1}{n} \sum_{i=1}^{n}(Y_{i}-\hat{Y_{i}})^{2}
	$$
	\item Mean Absolute Error (MAE):
	$$
	M A E=\frac{1}{n} \sum_{i=1}^{n}|Y_{i}-\hat{Y_{i}}|
	$$
\end{itemize}
In the short series prediction, we employ Root Relative Squared Error (RSE) and Empirical Correlation Coefficient (CORR) as indicators and follow the setting~\cite{c:33}. The formula is as follows:

\begin{itemize}
	\item Root Relative Squared Error (RSE):
	$$
	R S E=\frac{\sqrt{\sum_{t, i}\left(Y_{t i}-\hat{Y}_{t i}\right)^{2}}}{\sqrt{\sum_{t, i}\left(Y_{t i}-\overline{Y}\right)^{2}}}
	$$
	\item Empirical Correlation Coefficient (CORR):
	$$
	C O R R~(Y,\hat{Y})=\frac{1}{n} \sum_{i=1}^{n} \frac{\sum_{t}\left(Y_{t i}-\overline{Y_{i}}\right)\left(\hat{Y}_{t i}-\overline{\hat{Y}_{i}}\right)}{\sqrt{\sum_{t}\left(Y_{t i}-\overline{Y_{i}}\right)^{2}\left(\hat{Y}_{t i}-\overline{\hat{Y}_{i}}\right)^{2}}}
	$$
\end{itemize}

\section{Reproducibility Checklist}

\begin{itemize}
	\item [1.] This paper...
	\begin{itemize}
		\item Includes a conceptual outline and/or pseudocode description of AI methods introduced (yes)
		\item Clearly delineates statements that are opinions, hypothesis, and speculation from objective facts and results (yes)
		\item Provides well marked pedagogical references for less-familiare readers to gain background necessary to replicate the paper (yes)
	\end{itemize}     
	\item [2.] Does this paper make theoretical contributions? (yes)
	\begin{itemize}
		\item All assumptions and restrictions are stated clearly and formally. (partial)
		\item All novel claims are stated formally (e.g., in theorem statements). (partial)
		\item Proofs of all novel claims are included. (partial)
		\item Proof sketches or intuitions are given for complex and/or novel results. (yes)
		\item Appropriate citations to theoretical tools used are given. (partial)
		\item All theoretical claims are demonstrated empirically to hold. (partial)
		\item All experimental code used to eliminate or disprove claims is included. (yes)
	\end{itemize}     
	\item [3.] Does this paper rely on one or more datasets? (yes)
	\begin{itemize}
		\item A motivation is given for why the experiments are conducted on the selected datasets. (yes)
		\item All novel datasets introduced in this paper are included in a data appendix. (yes)
		\item All novel datasets introduced in this paper will be made publicly available upon publication of the paper with a license that allows free usage for research purposes. (yes)
		\item All novel datasets introduced in this paper will be made publicly available upon publication of the paper with a license that allows free usage for research purposes. (yes)
		\item All datasets drawn from the existing literature (potentially including authors’ own previously published work) are publicly available. (yes)
		\item All datasets that are not publicly available are described in detail, with explanation why publicly available alternatives are not scientifically satisficing. (yes)
	\end{itemize} 
\end{itemize}

\begin{table*}[t]
	\centering
	\begin{tabular}{|c|c|cc|cc|cc|cc|cc|}
		\hline
		\multicolumn{1}{|c|}{\multirow{3}[6]{*}{Methods}} & \multirow{3}[6]{*}{Metrics} & \multicolumn{2}{c|}{ETTh1} & \multicolumn{2}{c|}{ETTh2} & \multicolumn{2}{c|}{ETTm1} & \multicolumn{2}{c|}{Weather} & \multicolumn{2}{c|}{Electricity} \bigstrut\\
		\cline{3-12}          & \multicolumn{1}{c|}{} & \multicolumn{2}{c|}{horizon} & \multicolumn{2}{c|}{horizon} & \multicolumn{2}{c|}{horizon} & \multicolumn{2}{c|}{horizon} & \multicolumn{2}{c|}{horizon} \bigstrut\\
		\cline{3-12}          & \multicolumn{1}{c|}{} & 24    & 48    & 24    & 48    & 24    & 48    & 24    & 48    & 48    & 168 \bigstrut\\
		\hline
		\multicolumn{1}{|c|}{\multirow{2}[2]{*}{LogTrans}} & MSE   & 0.656  & 0.670  & 0.726  & 1.728  & 0.341  & 0.495  & 0.365  & 0.496  & 0.267  & 0.290  \bigstrut[t]\\
		& MAE   & 0.600  & 0.611  & 0.638  & 0.944  & 0.495  & 0.527  & 0.405  & 0.485  & 0.366  & 0.382  \bigstrut[b]\\
		\hline
		\multicolumn{1}{|c|}{\multirow{2}[2]{*}{Reformer}} & MSE   & 0.887  & 1.159  & 1.381  & 1.715  & 0.598  & 0.952  & 0.583  & 0.633  & 1.312  & 1.453  \bigstrut[t]\\
		& MAE   & 0.630  & 0.750  & 1.475  & 1.585  & 0.489  & 0.645  & 0.497  & 0.556  & 0.911  & 0.975  \bigstrut[b]\\
		\hline
		\multicolumn{1}{|c|}{\multirow{2}[2]{*}{LSTMa}} & MSE   & 0.536  & 0.616  & 1.049  & 1.331  & 0.511  & 1.280  & 0.476  & 0.763  & 0.388  & 0.492  \bigstrut[t]\\
		& MAE   & 0.528  & 0.577  & 0.689  & 0.805  & 0.517  & 0.819  & 0.464  & 0.589  & 0.444  & 0.498  \bigstrut[b]\\
		\hline
		\multicolumn{1}{|c|}{\multirow{2}[2]{*}{LSTNet}} & MSE   & 1.175  & 1.344  & 2.632  & 3.487  & 1.856  & 1.909  & 0.575  & 0.622  & 0.279  & 0.318  \bigstrut[t]\\
		& MAE   & 0.793  & 0.864  & 1.337  & 1.577  & 1.058  & 1.085  & 0.507  & 0.553  & 0.337  & 0.368  \bigstrut[b]\\
		\hline
		\multicolumn{1}{|c|}{\multirow{2}[2]{*}{Informer}} & MSE   & 0.509  & 0.551  & 0.446  & 0.934  & 0.325  & 0.472  & 0.353  & 0.464  & 0.269  & 0.300  \bigstrut[t]\\
		& MAE   & 0.523  & 0.563  & 0.523  & 0.733  & 0.440  & 0.537  & 0.381  & 0.455  & 0.351  & 0.376  \bigstrut[b]\\
		\hline
		\multicolumn{1}{|c|}{\multirow{2}[2]{*}{Autoformer}} & MSE   & 0.408  & 0.443  & 0.302  & 0.364  & 0.150  & 0.216  & \textbf{0.175 } & \textbf{0.224 } & 0.183  & 0.210  \bigstrut[t]\\
		& MAE   & 0.434  & 0.451  & 0.374  & 0.417  & 0.264  & 0.315  & \textbf{0.259 } & \textbf{0.305 } & 0.299  & 0.325  \bigstrut[b]\\
		\hline
		\multicolumn{1}{|c|}{\multirow{2}[2]{*}{SCINet}} & MSE   & 0.353  & 0.389  & \textit{0.188 } & 0.339  & \textit{0.128 } & \textit{0.157 } & \textit{0.322 } & 0.421  & \textit{0.151 } & 0.171  \bigstrut[t]\\
		& \multicolumn{1}{c|}{MAE} & 0.385  & 0.411  & 0.287  & 0.400  & 0.231  & 0.265  & \textit{0.346}  & 0.431  & \textit{0.252 } & 0.275  \bigstrut[b]\\
		\hline
		\multicolumn{1}{|c|}{\multirow{2}[2]{*}{LightTS}} & MSE   & \textbf{0.314 } & \textit{0.355 } & \textbf{0.178 } & \textit{0.251 } & \textbf{0.105 } & \textbf{0.139 } & 0.326  & \textit{0.387 } & \textbf{0.140 } & \textbf{0.150 } \bigstrut[t]\\
		& MAE   & \textbf{0.356 } & \textit{0.384 } & \textit{0.269 } & \textit{0.326 } & \textbf{0.197 } & \textbf{0.235 } & 0.351  & \textit{0.402 } & \textbf{0.244 } & \textbf{0.254 } \bigstrut[b]\\
		\hline
		\multicolumn{1}{|c|}{\multirow{2}[2]{*}{PureTS}} & MSE   & \textit{0.352 } & \textbf{0.350 } & 0.191  & \textbf{0.237 } & 0.137  & 0.169  & 0.406  & 0.466  & 0.221  & \textit{0.159 } \bigstrut[t]\\
		& MAE   & \textit{0.369 } & \textbf{0.367 } & \textbf{0.268 } & \textbf{0.299 } & \textit{0.229 } & \textit{0.263}  & 0.396  & 0.449  & 0.274  & \textit{0.250 } \bigstrut[b]\\
		\hline
	\end{tabular}%
	\caption{Supplementary results for long sequence prediction tasks. The model with the best result is in \textbf{bold}, the second result in \textit{italics}. The experimental results except PureTS are from~\cite{zhang2022less}.}
	\label{tableadd2}%
\end{table*}%

\begin{table*}[htbp]
	\centering
	\small
	
	\begin{tabular}{|c|c|cccc|cccc|}
		\hline
		\multicolumn{1}{|c|}{\multirow{3}[6]{*}{Methods}} & \multirow{3}[6]{*}{Metrics} & \multicolumn{4}{c|}{Traffic} & \multicolumn{4}{c|}{Electricity} \bigstrut\\
		\cline{3-10}          & \multicolumn{1}{c|}{} & \multicolumn{4}{c|}{horizon} & \multicolumn{4}{c|}{horizon} \bigstrut\\
		\cline{3-10}          & \multicolumn{1}{c|}{} & 3     & 6     & 12    & 24    & 3     & 6     & 12    & 24 \bigstrut\\
		\hline
		\multicolumn{1}{|c|}{\multirow{2}[2]{*}{AR}} & RSE   & 0.5991 & 0.6218 & 0.6252 & 0.6300 & 0.0995 & 0.1035 & 0.1050 & 0.1054 \bigstrut[t]\\
		& CORR  & 0.7752 & 0.7568 & 0.7544 & 0.7591 & 0.8845 & 0.8632 & 0.8691 & 0.8595 \bigstrut[b]\\
		\hline
		\multicolumn{1}{|c|}{\multirow{2}[2]{*}{VARMLP}} & RSE   & 0.5582 & 0.6579 & 0.6023 & 0.6146 & 0.1393 & 0.1620 & 0.1557 & 0.1274 \bigstrut[t]\\
		& CORR  & 0.8245 & 0.7695 & 0.7929 & 0.7891 & 0.8708 & 0.8389 & 0.8192 & 0.8679 \bigstrut[b]\\
		\hline
		\multicolumn{1}{|c|}{\multirow{2}[2]{*}{GP}} & RSE   & 0.6082 & 0.6772 & 0.6406 & 0.5995 & 0.1500 & 0.1907 & 0.1621 & 0.1273 \bigstrut[t]\\
		& CORR  & 0.7831 & 0.7406 & 0.7671 & 0.7909 & 0.8670 & 0.8334 & 0.8394 & 0.8818 \bigstrut[b]\\
		\hline
		\multicolumn{1}{|c|}{\multirow{2}[2]{*}{RNN-GRU}} & RSE   & 0.5358 & 0.5522 & 0.5562 & 0.5633 & 0.1102 & 0.1144 & 0.1183 & 0.1295 \bigstrut[t]\\
		& CORR  & 0.8511 & 0.8405 & 0.8345 & 0.8300 & 0.8597 & 0.8623 & 0.8472 & 0.8651 \bigstrut[b]\\
		\hline
		\multicolumn{1}{|c|}{\multirow{2}[2]{*}{TCN}} & RSE   & 0.5459 & 0.6061 & 0.6367 & 0.6586 & 0.0892 & 0.0974 & 0.1053 & 0.1091 \bigstrut[t]\\
		& CORR  & 0.8486 & 0.8205 & 0.8048 & 0.7921 & 0.9232 & 0.9121 & 0.9017 & 0.9101 \bigstrut[b]\\
		\hline
		\multicolumn{1}{|c|}{\multirow{2}[2]{*}{MTGNN}} & RSE   & \textit{0.4162} & 0.4754 & \textit{0.4461} & 0.4535 & \textbf{0.0745} & 0.0878 & \textbf{0.0916} & \textbf{0.0953} \bigstrut[t]\\
		& CORR  & \textbf{0.8963} & 0.8667 & \textbf{0.8794} & \textbf{0.8810} & \textit{0.9474} & \textit{0.9316} & \textit{0.9278} & \textit{0.9234} \bigstrut[b]\\
		\hline
		\multicolumn{1}{|c|}{\multirow{2}[2]{*}{SCINet}} & RSE   & 0.4203 & \textit{0.4447} & 0.4536 & \textit{0.4477} & \textit{0.0758} & \textbf{0.0852} & \textit{0.0934} & \textit{0.0973} \bigstrut[t]\\
		& CORR  & \textit{0.8931} & \textbf{0.8802} & \textit{0.8760} & \textit{0.8783} & \textbf{0.9493} & \textbf{0.9386} & \textbf{0.9296} & \textbf{0.9272} \bigstrut[b]\\
		\hline
		\multicolumn{1}{|c|}{\multirow{2}[2]{*}{LightTS}} & RSE   & \textbf{0.3973} & \textbf{0.4335} & \textbf{0.4403} & \textbf{0.4416} & 0.0762 & \textit{0.0876} & 0.0935 & 0.0985 \bigstrut[t]\\
		& CORR  & 0.8900 & \textit{0.8731} & 0.8696 & 0.8699 & 0.9432 & 0.9304 & 0.9238 & 0.9191 \bigstrut[b]\\
		\hline
		\multicolumn{1}{|c|}{\multirow{2}[2]{*}{PureTS}} & RSE   & 0.4848 & 0.4951 & 0.4971 & 0.4985 & 0.0855 & 0.0969 & 0.1015 & 0.1036 \bigstrut[t]\\
		& CORR  & 0.8561 & 0.8492 & 0.8477 & 0.8467 & 0.9362 & 0.9156 & 0.9037 & 0.9052 \bigstrut[b]\\
		\hline
		\multicolumn{1}{|c|}{\multirow{2}[2]{*}{PureTS\_S}} & RSE   & 0.5267 & 0.5511 & 0.5535 & 0.5471 & 0.1061 & 0.1139 & 0.1206 & 0.1141 \bigstrut[t]\\
		& CORR  & 0.8463 & 0.8335 & 0.8341 & 0.8372 & 0.9244 & 0.9039 & 0.8943 & 0.8914 \bigstrut[b]\\
		\hline
	\end{tabular}%
	\caption{Supplementary results for short sequence prediction tasks. The model with the best result is in \textbf{bold}, the second result in \textit{italics}. The experimental results except PureTS are from~\cite{zhang2022less}.}
	\label{table3}%
\end{table*}%

\end{document}